\documentclass[10pt,twocolumn,letterpaper]{article}

\usepackage[pagenumbers]{cvpr} 
\usepackage{algorithm} 
\usepackage{algpseudocode}
\usepackage{booktabs}
\usepackage{threeparttable}
\usepackage[table]{xcolor}
\definecolor{linkblue}{rgb}{0.0,0.45,0.74}
\usepackage{subcaption}
\usepackage{pgfplots}
\pgfplotsset{compat=newest}
\usepgfplotslibrary{polar}

\pgfplotscreateplotcyclelist{pastel}{
{blue!35!white,   mark=*},
{red!35!white,    mark=square*},
{orange!35!white, mark=triangle*},
{black!35,        mark=diamond*}
}

\usetikzlibrary{fillbetween}
\usepackage[utf8]{inputenc}
\usepackage{geometry}
\usepackage{xcolor}
\usepackage{tcolorbox}
\usepackage{enumitem}
\usepackage{utfsym}

\usepackage{pifont}      
\usepackage{multirow}

\usepackage{siunitx}
\newcommand{\cmark}{\textcolor{green!60!black}{\ding{51}}}
\newcommand{\xmark}{\textcolor{red!70!black}{\ding{55}}}

\usepackage{dblfloatfix}
\definecolor{cvprblue}{rgb}{0.21,0.49,0.74}
\usepackage[pagebackref,breaklinks,colorlinks,allcolors=cvprblue]{hyperref}

\def\paperID{5205} 
\def\confName{CVPR}
\def\confYear{2026}

\title{Resolving Evidence Sparsity: Agentic Context Engineering for Long-Document Understanding}

\author{Keliang Liu$^{1}$ $\quad$
        Zizhi Chen$^{1}$ $\quad$
        Mingcheng Li$^{1}\quad$
        Jingqun Tang$^{3}\quad$ \\
        Dingkang Yang$^{1,2\,\dagger\,\usym{1F396}}$ $\quad$ 
        Lihua Zhang$^{1,2\,\dagger}$ \\ 
         \small $^\dagger$corresponding authors $\quad$ $^{\usym{1F396}}$project lead  \\
        $^1$College of Intelligent Robotics and Advanced Manufacturing, Fudan University\\
        $^2$Fysics Intelligence Technologies Co., Ltd. (Fysics AI) \\
        $^3$ByteDance \\
{\tt\small klliu25@m.fudan.edu.cn,\, \{dkyang20,\,lihuazhang\}@fudan.edu.cn}
}

\begin{document}
\maketitle

\begin{abstract}
\frenchspacing
Document understanding is a long-standing practical task. Vision-Language Models (VLMs) have gradually become a primary approach in this domain, demonstrating effective performance on single-page tasks. However, their effectiveness diminishes when handling long documents. In such scenarios, clues are often scattered across multiple pages and modalities, and redundancy from lengthy inputs can impair the model's judgment. While retrieval-augmented generation mitigates this issue by filtering for question-relevant content, the retrieved results still contain substantial redundancy. To address these limitations, we propose SLEUTH, a multi-agent framework. Concretely, SLEUTH orchestrates a retriever and four collaborative agents in a coarse-to-fine process. The framework identifies key textual and visual clues within the retrieved pages, filters for salient visual evidence such as tables and charts, and analyzes the query to devise a reasoning strategy. It ultimately synthesizes a distilled, evidence-dense multimodal context to generate the final prediction. SLEUTH is model-agnostic and scalable. When paired with advanced VLM backbones, it consistently improves performance on multiple long-document benchmarks, achieving SOTA results. Ablation studies verify each module’s effectiveness and confirm the benefits of our hierarchical refinement paradigm.
\end{abstract}

\section{Introduction}
\label{sec:intro}
Documents are one of the fundamental forms of human information preservation and transmission~\cite{ma2024mmlongbench, chia2025m}. Understanding documents with complex layouts and multimodal components has long been a pragmatic challenge and also serves as a benchmark for evaluating the multimodal long-context reasoning capability of AI systems~\cite{deng-etal-2025-longdocurl, duan2025docopilot, wang-etal-2024-docllm}. Documents convey rich visual information, not only through textual content but also via charts, page layouts, tables, and even fonts. Traditional document question answering methods usually employ OCR-based pipelines~\cite{wang2024mineru, wei2024general} to extract textual content and then feed it into large language models (LLMs) for response generation. However, such approaches often lose critical multimodal cues, leading to imperfect and shallow document comprehension. Recently, Multimodal Large Language Models (MLLMs)~\cite{li2024llava, Qwen2.5-VL, wang2025internvl3_5, vteam2025glm45vglm41vthinkingversatilemultimodal, hurst2024gpt, anthropic2025claude} have achieved remarkable progress in document understanding tasks owing to their intrinsic multimodal capabilities, particularly in single-page document understanding~\cite{mathew2021docvqa, masry2022chartqa, mathew2022infographicvqa, zhu2022towards}. Nevertheless, their ability to comprehend long-context documents remains uncertain~\cite{zhuang2025self, zhou2024rethinking, wang2025multimodal}. Long documents introduce extensive contexts where most information is redundant, while the key evidence required to answer a query is often sparse and scattered across multiple pages and modalities. This characteristic poses substantial challenges to existing VLMs. The core objective of long-document understanding lies in precisely locating informative evidence from massive content and organizing it into a high-quality contextual representation suitable for reasoning and response generation.
\begin{figure*}[!htbp]
  \centering
  \includegraphics[width=\linewidth]{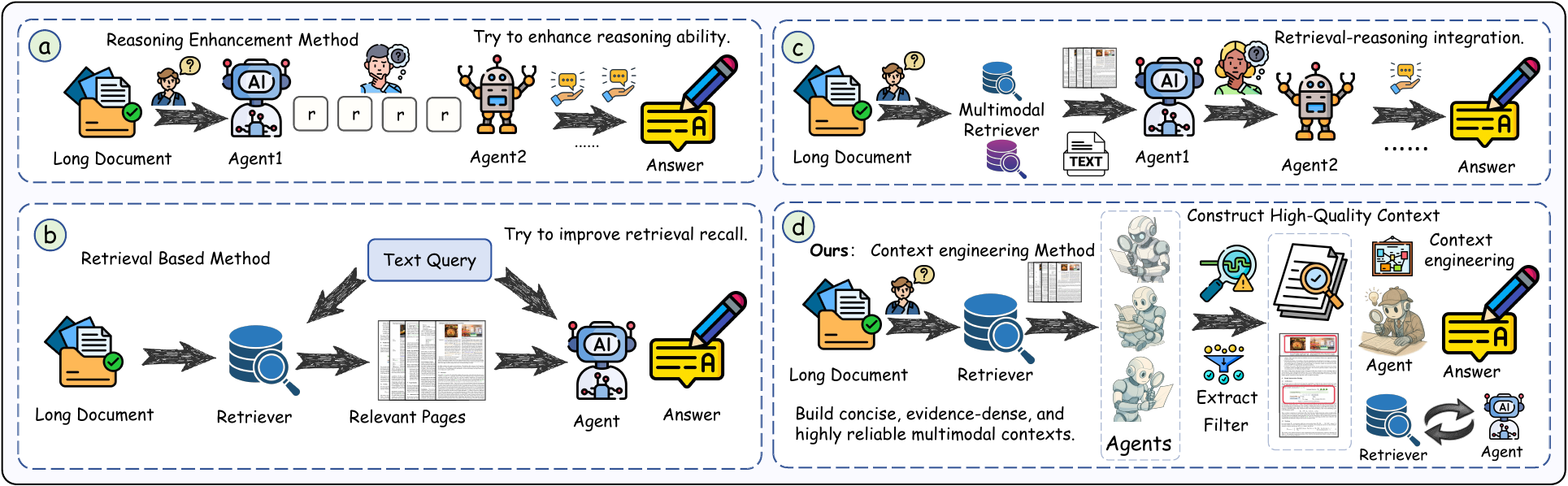}
  \caption{Comparison with mainstream methods. (a) Strengthening reasoning via agent optimization; 
(b) Improving recall through retrieval augmentation; 
(c) Combining (a) and (b); 
(d) Our method focuses on constructing evidence-dense contexts.
}
  \label{fig:intro}
  \vspace{-7pt}
\end{figure*} 

Recent document understanding studies mainly follow three paths, namely enhancing agent reasoning, improving retrieval recall, and combining the two.
As shown in Figure~\ref{fig:intro}\textcolor{linkblue}{a}, multi-agent approaches such as MACT~\cite{yu2025visual} strengthen reasoning through agent collaboration and training, and achieve good results in long-document understanding. However, when answering questions, these methods still take long contexts as input, with large amounts of redundant information that interferes with reasoning. Figure~\ref{fig:intro}\textcolor{linkblue}{b} demonstrates that Retrieval-Augmented Generation (RAG) methods~\cite{faysse2024colpali, cho2024m3docrag, chen2025svrag, wang2025vidorag, wu2025molorag} retrieve document content relevant to the query and achieve much better performance than directly reading the entire long document. However, the retrieved results still contain much irrelevant information, and VLMs have difficulty identifying the few key pieces of evidence scattered across multiple pages. Refer to Figure~\ref{fig:intro}\textcolor{linkblue}{c}, methods such as MDocAgent~\cite{han2025mdocagent} combine RAG with multi-agent guided thinking, achieving performance gains. However, the reasoning process still involves long and noisy contexts. Our method is illustrated in Figure~\ref{fig:intro}\textcolor{linkblue}{d}. From the perspective of context engineering, we introduce \underline{S}equential \underline{L}ong-document \underline{E}vidence \underline{U}ncovering through mul\underline{T}i-agent \underline{H}ierarchical refinement (SLEUTH), a plug-and-play multi-agent framework that, in a training-free manner, builds concise and evidence-dense contexts from noisy long-context retrieval results, thereby efficiently improving document understanding performance. A standard retriever first narrows the search space; four collaborative agents then mine page-level textual and visual clues, filter irrelevant page images, and adapt the reasoning strategy to the query’s difficulty before producing the final answer. By operating page-wise, SLEUTH keeps the effective context length fixed, enabling accuracy to scale with larger retrieval top-K without amplifying hallucinations. Our method can also be combined with other approaches to further improve document understanding capabilities.

The core contributions of this work are summarized as follows: (i) We propose SLEUTH, a training-free multi-agent framework that efficiently mines key evidence from noisy long documents and constructs evidence-dense multimodal contexts. (ii) We design two complementary agents: the Clue Discovery Agent and the Page Screening Agent, responsible for clue extraction and visual filtering, respectively. Their collaboration enables a structured and refined contextual representation. (iii) We introduce an evidence and difficulty-aware mechanism that allows the system to adaptively select reasoning strategies based on query complexity and evidence context, leading to consistent performance improvements across multiple long-document understanding benchmarks.

\noindent To the best of our knowledge, this is the first work that investigates long-document understanding from the perspective of constructing concise, evidence‑dense contexts. Extensive experiments on four benchmarks, including MMLongBench~\cite{ma2024mmlongbench}, LongDocURL~\cite{deng-etal-2025-longdocurl}, PaperTab~\cite{hui2024uda}, and FetaTab~\cite{hui2024uda}, demonstrate that our approach is model-agnostic and achieves SOTA performance across different VLM backbones. Furthermore, ablation studies verify the effectiveness of each agent and emphasize the importance of constructing compact, trustworthy, and evidence-dense contexts for robust long-document understanding. We believe this method, together with reinforcement reasoning and retrieval-based methods, will jointly advance the progress in the field of long-document understanding.

\section{Related Work}
\label{sec:related}

\textbf{Document Analysis and Understanding.} 
Document Analysis and Understanding, a long-standing task with strong practical needs~\cite{ma2024mmlongbench, chia2025m}, has gained renewed attention with the rise of multimodal large language models~\cite{hurst2024gpt, anthropic2025claude, comanici2025gemini, Qwen2.5-VL, wang2025internvl3_5, vteam2025glm45vglm41vthinkingversatilemultimodal} These models show strong performance on benchmarks like DocVQA~\cite{mathew2021docvqa}, ChartQA~\cite{masry2022chartqa}, InfoVGA~\cite{mathew2022infographicvqa}, and TAT-DQA~\cite{zhu2022towards}. However, real-world documents are often long and complex, spanning text, tables, charts, and images. Consequently, research has shifted toward long-document settings that demand cross-page and multimodal reasoning~\cite{ma2024mmlongbench, deng-etal-2025-longdocurl, hui2024uda, van2023document, cho2024m3docrag, chia2025m, duan2025docopilot}. Traditional OCR-based LLM methods can only process text, whereas VLM-based approaches directly interpret document images via multimodal capabilities and have become the mainstream paradigm. DeepSeek-OCR \cite{wei2025deepseekocrcontextsopticalcompression} shows that unified visual mappings can compress contexts while retaining layout cues. We adopt a related intuition by representing pages visually, which preserves structure and supports fine-grained perception and reasoning.\\
\textbf{Context Engineering.} 
Benefiting from the development of learning-based approaches~\cite{yang2025improvingmsa,yang2024pediatricsgpt,yang2025improving,yang2025medaide,liu2025reinforcement,lin2025sail,yang2023target,yang2022learning,yang2022disentangled,yang2022emotion,yang2023context,yang2024robust,yang2024asynchronous,yang2024towards,yang2024MCIS}, context engineering involves the construction of dynamic systems that furnish accurate information and tools in appropriate formats, thereby enabling LLMs to effectively execute tasks. Suboptimal agent performance is frequently attributable to the inadequate provision of context, instructions, and tools to the model. Numerous studies~\cite{zhuang2025self, chen2025svrag, ma2024mmlongbench, zhou2024rethinking, wang2025multimodal, wang-etal-2024-docllm} reveal that, despite the ultra-long context windows of MLLMs, performance degrades sharply with increasing context length. The cornerstone of effective document analysis and understanding lies in high-quality context engineering. Retrieval-based methods like Colpali~\cite{faysse2024colpali}, Colbert~\cite{khattab2020colbert}, and BGE M3~\cite{chen2024bge} support document embeddings, while RAG frameworks such as M3DocRAG~\cite{cho2024m3docrag}, SV-RAG~\cite{chen2025svrag}, ViDoRAG~\cite{wang2025vidorag} and MoLoRAG~\cite{wu2025molorag} retrieve relevant pages to reduce context length. Dynamic methods like Doc-React~\cite{wu2025doc} refine sub-queries, and ACE~\cite{zhang2025agentic} or Sun \textit{et al.}~\cite{sun2025scaling} advocate proactive context management. Multi-agent systems such as CoA~\cite{zhang2024chain} and MDocAgent~\cite{han2025mdocagent} distribute reasoning across agents, while reinforcement learning (RL) approaches (VRAG-RL~\cite{wang2025vrag}, MACT~\cite{yu2025visual}, ReMemR1~\cite{shi2025look}, CARE~\cite{wang2025improving}) enhance evidence gathering and retrieval. However, the above methods do not optimize for the fact that MLLM performance degrades as context length increases. By dynamically recording clues and filtering visual information, we construct concise yet information-dense contexts from overly long and redundant retrieval results, thereby enhancing understanding capability.

\section{Methodology}
\label{sec:method}

\subsection{Overall Framework}

\begin{figure*}[!htbp]
  \centering
  \includegraphics[width=\linewidth]{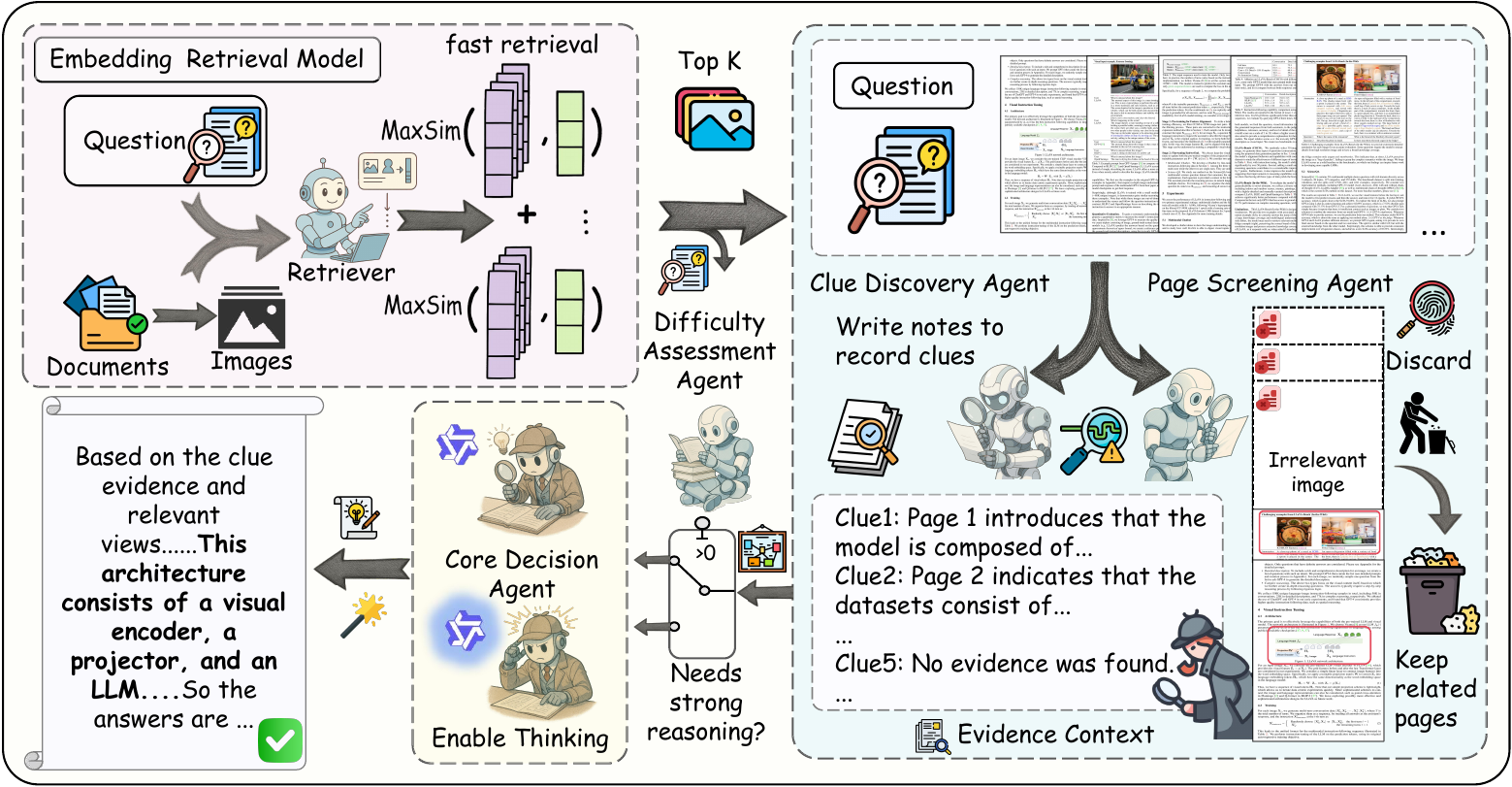}
\caption{\textbf{Overall Framework.} 
SLEUTH adopts a coarse-to-fine pipeline: (1) a visual retriever selects Top-$K$ pages; (2) Clue Discovery Agent records and refines evidence, Page Screening Agent filters irrelevant page images, (3) Difficulty Assessment Agent analyzes query complexity, and (4) Core Decision Agent reasons over the distilled, evidence-dense context.}
  \label{fig:overview}
  \vspace{-7pt}
\end{figure*}

As shown in Figure \ref{fig:overview}, given a question $Q$ and a document $D=\{p_1,p_2,\dots,p_N\}$, where each $p_i$ denotes an individual page represented as an RGB image and $N$ is the total number of pages, the goal of long-document question answering is to generate an accurate answer $A$ to $Q$ based on the evidence contained in $D$. Our SLEUTH
is a training-free and plug-and-play framework with the following workflow: First, a standard retrieval model narrows down the evidence search space, reducing computational cost. The retrieved coarse-grained relevant pages are then processed by our multi-agent system. Leveraging the strong single-image reasoning capability of VLMs, Clue Discovery Agent scans each retrieved page and records structured textual and visual cues. In parallel, Page Screening Agent selects page images containing task-relevant tables, charts, or diagrams and filters out irrelevant ones. Difficulty Assessment Agent analyzes the query content, generates strategic instructions, and guides Core Decision Agent to adopt the optimal reasoning mode. This process produces a denoised multimodal context that is both highly reliable and rich in evidential clues. Finally, Core Decision Agent infers the final answer based on the query and the refined high-confidence context. 

\subsection{Coarse-grained Visual Retrieval}
\label{subsec:retrieval}
To save computation and time, we first efficiently localize a small set of pages most relevant to the query $Q$ across the entire long document. We use Colpali~\cite{faysse2024colpali} to encode the question $Q$ into a sequence of textual embeddings ${\mathbf{q}t}{t=1}^{n_Q}$ and each page image $p_i$ into a set of visual embeddings ${\mathbf{v}{i,j}}{j=1}^{n_i}$. The page-level relevance is then defined as:

\begin{equation}
\label{eq:li}
s_i \;=\; \sum_{t=1}^{n_Q}\;\max_{1\le j\le n_i}\;\langle \mathbf{q}_t, \mathbf{v}_{i,j}\rangle .
\end{equation}
Based on the relevance scores $\{s_i\}_{i=1}^N$, we select the top-$K$ indices $\mathcal{I}_K$ to obtain the set of candidate pages:
\begin{equation}
\label{eq:topk}
P_K \;=\; \{\, p_i \mid i \in \mathrm{TopK}(\{s_i\}_{i=1}^{N}) \,\}.
\end{equation}
The coarse-grained retrieval stage performs page-level selection in a purely visual manner, ensuring that $K\!\ll\!N$ and thus significantly reduces computation time.

\subsection{Evidence Refinement and Visual Screening}
\label{subsec:stage2}
After page-level retrieval, the candidate set still contains a large amount of redundant information. The system must further identify the specific parts that truly contain critical evidence. In SLEUTH, we introduce two complementary agents: Clue Discovery Agent and Page Screening Agent. These two agents operate in parallel—the former focuses on discovering and recording interpretable structured clues, while the latter analyzes the overall visual composition of each page and filters out irrelevant images. Through this collaborative mechanism, the system constructs an evidence-dense multimodal context, providing a solid foundation for the final reasoning stage.

Clue Discovery Agent treats each page as the minimal processing unit. The MLLMs sequentially process each candidate page image and extract query-relevant evidence. For any $p_i \in P_K$, the agent identifies potential clues at the region level such as text lines, table cells, or chart areas and produces a structured clue set as follows:
\[
\mathcal{E}_i=\{e_{i,1},e_{i,2},\dots,e_{i,M_i}\}.
\]
Each evidence unit is organized as a structured record:
\begin{align*}
\hfill
e_{i,m} = \bigl(&\mathrm{page}=i,\;
                \mathrm{region}=\mathbf{w}_{i,m},\;
                \mathrm{content}=c_{i,m},\\
               &\mathrm{insight}=k_{i,m},\;
                \mathrm{rationale}=r_{i,m}\bigr),
\end{align*}
where $\mathbf{w}_{i,m}$ denotes the spatial position or cell coordinates of the clue, \textit{Content} $c_{i,m}$ represents the clue extracted from $p_i$, $k_{i,m}$ denotes the key insight inferred by the agent based on the clue, and $r_{i,m}$ records the complete chain-of-thought (CoT) reasoning trace for relevance analysis. This design ensures that each clue not only carries semantic content but also preserves its source, spatial location, and adoption rationale, facilitating provenance tracking and interpretability verification during subsequent reasoning stages.

Although Clue Discovery Agent captures fine-grained clues within each page, for pages containing complex visual elements such as tables, charts, or embedded illustrations the textual records alone may still be insufficient to reflect all informative cues. To compensate for this limitation, Page Screening Agent also analyzes each candidate page in $P_K$, focusing on the visual information and its semantic alignment with the query intent. Let the set of salient visual elements on the $i$-th page be defined as:
\[
\mathcal{V}_i=\{v_{i,1},v_{i,2},\dots,v_{i,K_i}\},
\]
Each visual element $v_{i,k}$ represents a table, chart, or key illustration within the page. Page Screening Agent performs joint semantic–visual reasoning over the entire page and outputs both the relevance level between the page and the query, as well as the corresponding reasoning explanation.  
Specifically, for a candidate page $p_i$, the agent generates a discrete relevance label based on the holistic page content:
\[
y_i \in \{\text{CR},\,\text{R},\,\text{IR}\},
\]
and simultaneously provides an explanatory description $r_i^{\mathrm{page}}$ that clarifies the reasoning process behind its decision. CR, R, and IR refer to Completely Relevant, Relevant, and Irrelevant, respectively. 
This classification result is entirely derived from the model’s own multimodal reasoning: the agent integrates the page’s layout structure, table organization, chart content, and textual semantics to infer the logical relationships between these visual elements and the query, thereby determining whether the page should be retained. The number of retained pages adapts dynamically according to the query. For example, on LongDocURL~\cite{deng-etal-2025-longdocurl}, when retrieving the top-5 page images, only an average of 2.1 relevant visual pages are preserved. The final retained set is defined as:

\[
\widetilde{P} = \{\,p_i \in P_K \mid y_i \in \{\text{CR}, \text{R}\}\,\}.
\]

The two agents are complementary at different levels: Clue Discovery Agent provides fine-grained and interpretable textual evidence, while Page Screening Agent ensures that critical visual elements are sufficiently preserved. Together, they collaboratively construct a multimodal context:
\begin{equation}
\small
\mathcal{C}=\Big(\widetilde{P},\;\mathcal{E}=\bigcup_{p_i\in P_K}\mathcal{E}_i\Big),
\label{eq:context}
\end{equation}
where $\widetilde{\mathcal{P}}$ denotes the set of retained page images, and $\mathcal{E}$ represents the collection of evidence extracted from the original top-$K$ pages. Through this parallel–complementary collaboration mechanism, SLEUTH achieves compact input while maximizing evidence density, thereby substantially enhancing the contextual quality and robustness of long-document question answering. In certain cases, modality absence may occur. For example, some questions may result in all page images being filtered out, leading to a lack of visual inputs, or the query itself may be irrelevant to the document, yielding an unanswerable case. Such situations are reasonable and naturally handled: depending on the presence or absence of visual modality in the constructed context, the system automatically adjusts the prompt template fed into Core Decision Agent, avoiding inconsistencies such as providing visual instructions without corresponding visual inputs. Algorithm~\ref{alg:unified} illustrates the overall process of multimodal context construction.

\begin{algorithm}[t]
\caption{Retrieval and Context Construction}
\label{alg:unified}
\begin{algorithmic}[1]
\Require Natural-language question $Q$; document $D=\{p_1,\dots,p_N\}$; parameter $K$; 
\Ensure Filtered pages $\widetilde{P}$; global evidence set $\mathcal{E}$
\State Encode the query: $\{\mathbf{q}_t\}_{t=1}^{n_Q} \leftarrow \mathrm{EncodeQuery}(Q)$
\For{$i=1$ to $N$} \Comment{Relevance scoring}
    \State $\{\mathbf{v}_{i,j}\}_{j=1}^{n_i} \leftarrow \mathrm{VisualEncode}(p_i)$
    \State $s_i \leftarrow \sum_{t=1}^{n_Q}\max_{1\le j\le n_i}\langle \mathbf{q}_t, \mathbf{v}_{i,j}\rangle$ \Comment{Eq.~\eqref{eq:li}}
\EndFor
\State $\mathcal{I}_K \leftarrow \mathrm{TopK}_K(\{s_i\}_{i=1}^{N})$; \quad $P_K \leftarrow \{p_i\mid i\in\mathcal{I}_K\}$ 
\State Initialize $\widetilde{P}\!\leftarrow\!\emptyset$, $\mathcal{E}\!\leftarrow\!\emptyset$
\For{each page $p_i\!\in\!P_K$} \Comment{Two agents run in parallel}
  \Statex \hspace{1.2em}\textit{(a) Clue Discovery Agent: clue extraction}
  \State $\mathcal{R}_i \leftarrow \mathrm{ProposeRegions}(p_i)$; \quad $\mathcal{E}_i \leftarrow \emptyset$
  \For{each region $\mathbf{w}\!\in\!\mathcal{R}_i$}
      \State $(c,k,r) \leftarrow \mathrm{ClueDiscovery}(Q,p_i,\mathbf{w})$
      \State $e \leftarrow (\mathrm{page}\!=\!i,\,\mathrm{region}\!=\!\mathbf{w},\,\mathrm{content}\!=\!c,\,\mathrm{insight}\!=\!k,\,\mathrm{rationale}\!=\!r)$
      \State $\mathcal{E}_i \leftarrow \mathcal{E}_i \cup \{e\}$
  \EndFor
  \Statex \hspace{1.2em}\textit{(b) Page Screening Agent: visual relevance filtering}
  \State Initialize page-level candidate set $\widetilde{P} \leftarrow \emptyset$
  \State $(y_i,\, r_i^{\mathrm{page}}) \leftarrow \mathrm{PageScreen}(Q, p_i)$

\If{$y_i\in\{\mathrm{CR},\,\mathrm{R}\}$}
\State$\widetilde{P} \leftarrow \widetilde{P} \cup \{p_i\}$
  \EndIf

\EndFor
\State \Return $\widetilde{P}, \mathcal{E}$
\end{algorithmic}
\end{algorithm}

\subsection{Evidence and Difficulty-Aware Decision}
\label{subsec:reasoning}

Different questions require different levels of reasoning.  
Some queries can be directly answered based on the aggregated evidence, while others demand integration across multiple pages or numerical reasoning. Difficulty Assessment Agent analyzes the query and selects the appropriate model type, providing guiding instructions for subsequent reasoning. Given a question $Q$ and a structured context $\mathcal{C}$, the agent first determines the task difficulty level $d \in \{0,1\}$ and outputs an instruction set $\Gamma_d$:
\begin{equation}
d=\arg\max_{c\in\{0,1\}}f(c;Q,\mathcal{C}), 
\Gamma_d=\Psi(Q,\mathcal{C},d),
\label{eq:difficulty}
\end{equation}
where $d=0$ indicates a ordinary mode and $d=1$ denotes a reasoning mode . The ordinary mode corresponds to general Instruct-type models and is suitable for most basic queries, while the reasoning mode corresponds to Thinking-type models (Some MLLMs have a thinking version or can be enabled with thinking capabilities) capable of multi-step inference across pages and performing numerical computation. The instruction set $\Gamma_d$ summarizes key reasoning cues, such as ``\textit{requires cross-page aggregation}'', ``\textit{involves table calculation}'' or ``\textit{needs trend comparison}'' which guide the downstream decision-making process. The core reasoning is executed by Core Decision Agent.  
\begin{algorithm}[t]
\caption{Reasoning and Decision-Making Based on Evidence Context and Difficulty Perception}
\label{alg:reason}
\begin{algorithmic}[1]
\Require Question $Q$; structured context $\mathcal{C}=(\widetilde{P},\mathcal{E})$; MLLMs $\Phi$
\Ensure Final answer $A^\star$; evidence reference table $\mathbb{S}$
\State \textbf{Step 1.} \textit{difficulty assessment and instruction extraction.}
\State $(d,\Gamma_d)\leftarrow\Phi_{\mathrm{assess}}(Q,\mathcal{C})$ 
\Comment{Single inference: output difficulty $d\!\in\!\{0,1\}$ and task instructions $\Gamma_d$}
\State \textbf{Step 2.} \textit{Select reasoning mode.}
\If{$d=0$} 
    \State Select the Instruct-type model.
\Else
    \State Select the Thinking-type model.
\EndIf
\State \textbf{Step 3.} \textit{Final reasoning and answer generation.}
\State $(A^\star,\mathbb{S})\leftarrow\Phi_{\mathrm{reason}}(Q,\mathcal{C},\Gamma_d)$ 
\Comment{Model produces answer and evidence references}
\State \Return $A^\star,\mathbb{S}$
\end{algorithmic}
\end{algorithm}

This agent receives the instruction set $\Gamma_d$ and the structured context $\mathcal{C}$ from Difficulty Assessment Agent, and then invokes the corresponding type of MLLM to generate the final answer:
\begin{equation}
A^\star=\Phi(Q,\mathcal{C},\Gamma_d),
\label{eq:decision}
\end{equation}
where $\Phi$ denotes multimodal large language models' unified reasoning function capable of switching between different backbone models. During answer generation, Core Decision Agent simultaneously produces an evidence reference table $\mathbb{S}=\{\text{page index}, \text{evidence content}, \text{evidence source}\}$, ensuring verifiability of the reasoning process.

Through this difficulty-aware model selection mechanism, SLEUTH dynamically adapts to queries of varying complexity while maintaining a training-free paradigm, effectively balancing efficiency and reasoning depth.  
Algorithm~\ref{alg:reason} illustrates the overall reasoning procedure based on multimodal evidence context and difficulty perception.

\section{Experiments}
\label{sec:Experiments}
\begin{table*}[t]
\centering
\setlength{\tabcolsep}{18pt}
\renewcommand{\arraystretch}{1.15}
\caption{Results on MMLongBench-Doc~\cite{ma2024mmlongbench}. Accuracy (\%, higher is better) across different content types. SLEUTH achieves the highest average value and the highest scores in multiple subtasks. \textbf{Bold} denotes the best in each column.}
\label{tab:mmldoc_results}
\resizebox{\textwidth}{!}{%
\begin{threeparttable}
\begin{tabular}{lccccccc}
\toprule
\textbf{Method} & \textbf{Chart} & \textbf{Table} & \textbf{Pure-text} & \textbf{Layout} & \textbf{Figure} & \textbf{None} & \textbf{Avg.} \\
\midrule
Direct & 34.37 & 37.57 & 46.13 & 54.93 & 37.03 & 55.61 & 42.71 \\
M3DocRAG~\cite{cho2024m3docrag} & 53.94 & 44.15 & 54.63 & 53.52& 41.51 & 48.64 & 45.90 \\
MoLoRAG~\cite{wu2025molorag} & \textbf{54.85} & 46.89 & 54.86 & 54.93 & 46.41 & 51.67 & 48.75 \\
MDocAgent~\cite{han2025mdocagent} & 52.97 & 45.98 & 53.81 & \textbf{56.39} & 45.52 & 49.71 & 47.82 \\
Base & 54.72 & 44.76 & 53.33 & 53.52 & 44.92 & 52.68 & 46.76 \\
\rowcolor{gray!10}
\textbf{SLEUTH (ours)} & 53.27 & \textbf{47.55} & \textbf{59.26} & 53.52 & \textbf{50.27} & \textbf{67.38} & \textbf{52.77} \\
\bottomrule
\end{tabular}
\end{threeparttable}
}
\end{table*}

\begin{table*}[t]
\centering
\caption{Results on LongDocURL~\cite{deng-etal-2025-longdocurl},
PaperTab~\cite{hui2024uda} and FetaTab~\cite{hui2024uda}. 
Accuracy (\%, higher is better). 
SLEUTH achieves the best overall results across all tasks and datasets. 
\textbf{Bold} denotes the best in each column.}
\label{tab:longdocurl_papertab_fetatab}
\resizebox{\textwidth}{!}{%
\setlength{\tabcolsep}{18pt}
\renewcommand{\arraystretch}{1.12}

\begin{tabular}{lcccccc}
\toprule
\multirow{2}{*}{\textbf{Method}} & \multicolumn{4}{c}{\textbf{LongDocURL}~\cite{deng-etal-2025-longdocurl}} &
\multirow{2}{*}{\textbf{PaperTab}~\cite{hui2024uda}} &
\multirow{2}{*}{\textbf{FetaTab}~\cite{hui2024uda}} \\
\cmidrule(lr){2-5}
 & \textbf{Locating} & \textbf{Understanding} & \textbf{Reasoning} & \textbf{Avg.} &  &  \\
\midrule
M3DocRAG~\cite{cho2024m3docrag}  & 45.46 & 60.07 & 52.37 & 54.59 & 39.11 & 64.08 \\
MoLoRAG~\cite{wu2025molorag}   & 49.32 & 63.35 & 52.73 & 57.57 & 42.59 & 69.41 \\
MDocAgent~\cite{han2025mdocagent} & 44.74 & 57.46 & 51.46 & 53.11 & 39.86 & 66.55 \\
Base      & 46.04 & 61.56 & 51.09 & 55.18 & 38.88 & 64.16 \\
\rowcolor{gray!10}
\textbf{SLEUTH (ours)} & \textbf{53.63} & \textbf{65.67} & \textbf{52.99} & \textbf{59.96} & \textbf{43.09} & \textbf{70.46} \\
\bottomrule
\end{tabular}
}
\end{table*}

\begin{table*}[t]
\centering
\caption{Ablation of SLEUTH variants across different base models on MMLongBench-Doc~\cite{ma2024mmlongbench} and 
LongDocURL~\cite{deng-etal-2025-longdocurl}. 
Accuracy (\%, higher is better). We progressively enable each agent, Clue Discovery Agent (C), Page Screening Agent (P), and Difficulty Assessment Agent (D)—and vary the retriever Top-$K$ (1/3/5). Performance improves consistently as the agent system becomes more complete, with Top-5 achieving the best overall average.
\textbf{Bold} indicates the best average (Avg.).}
\label{tab:sleuth_combined_ablation}
\resizebox{\textwidth}{!}{%
\small
\setlength{\tabcolsep}{5pt}
\renewcommand{\arraystretch}{1.15}
\begin{tabular}{
    p{2.1cm} p{2.3cm} ccc
    *{7}{S[table-format=2.2]} | *{4}{S[table-format=2.2]}
}
\toprule
\multirow{2}{*}{\textbf{Model}} & \multirow{2}{*}{\textbf{Variants}} &
\multicolumn{3}{c}{\textbf{Agent Configuration}} &
\multicolumn{7}{c|}{\textbf{MMLongBench-Doc}} &
\multicolumn{4}{c}{\textbf{LongDocURL}} \\
\cmidrule(lr){3-5} \cmidrule(lr){6-12} \cmidrule(lr){13-16}
 & & \textbf{Clue.} & \textbf{Page.} & \textbf{Diff.} &
 {\textbf{Chart}} & {\textbf{Table}} & {\textbf{Pure-text}} & {\textbf{Layout}} & {\textbf{Figure}} & {\textbf{None}} & {\textbf{Avg.}} &
 {\textbf{Loc.}} & {\textbf{Und.}} & {\textbf{Reas.}} & {\textbf{Avg.}} \\
\midrule
\multirow{6}{*}{\textbf{Qwen3-VL-8B}}
 & Base          & -- & -- & -- & \textbf{54.72} & 44.76 & 53.33 & 53.52 & 44.92 & 52.68 & 46.76 & 46.04 & 61.56 & 51.09 & 55.18 \\
 & SLEUTH (C)    & \cmark & \xmark & \xmark & 45.28 & 46.15 & 54.81 & \textbf{56.39} & 48.02 & \textbf{69.23} & 48.61 & 49.03 & 62.05 & 54.27 & 57.15 \\
 & SLEUTH (P)    & \cmark & \cmark & \xmark & 51.40 & 46.15 & 55.56 & 50.70 & 50.27 & 67.38 & 51.29 & 53.21 & 65.35 & 52.21 & 59.49 \\
 & SLEUTH (Top1) & \cmark & \cmark & \cmark & 41.46 & 40.41 & 49.33 & 45.07 & 38.24 & 74.25 & 44.92 & 52.69 & 58.03 & 36.69 & 52.88 \\
 & SLEUTH (Top3) & \cmark & \cmark & \cmark & 49.21 & 43.49 & 55.08 & 49.25 & 49.94 & 67.38 & 49.65 & \textbf{57.48} & 62.64 & 49.14 & 58.38 \\
\rowcolor{gray!10}
 & SLEUTH (Top5) & \cmark & \cmark & \cmark & 53.27 & \textbf{47.55} & \textbf{59.26} & 53.52 & \textbf{50.27} & 67.38 & \textbf{52.77} & 53.63 & \textbf{65.67} & \textbf{52.99} & \textbf{59.96} \\
\midrule
\multirow{6}{*}{\shortstack[l]{\textbf{GLM-4.1V-}\\\textbf{Thinking-8B}}}
 & Base          & -- & -- & -- & 51.01 & 39.07 & 46.67 & 49.25 & 41.37 & 94.82 & 53.05 & 48.03 & 64.40 & 54.04 & 57.75 \\
 & SLEUTH (C)    & \cmark & \xmark & \xmark & 45.85 & 41.88 & 49.52 & 53.52 & 42.27 & \textbf{95.12} & 53.22 & 51.24 & 63.36 & 53.37 & 58.08 \\
 & SLEUTH (Top1) & \cmark & \cmark & -- & 42.62 & 32.45 & 48.59 & 50.70 & 46.79 & 84.08 & 49.38 & 37.61 & 57.98 & 48.66 & 50.34 \\
 & SLEUTH (Top3) & \cmark & \cmark & -- & 50.65 & 39.93 & 57.26 & 54.93 & \textbf{47.38} & 79.10 & 54.34 & 45.19 & 64.64 & \textbf{55.40} & 57.29 \\
\rowcolor{gray!10}
 & SLEUTH (Top5) & \cmark & \cmark & -- & \textbf{52.38} & \textbf{44.77} & \textbf{57.34} & \textbf{59.15} & 47.29 & 94.62 & \textbf{57.93} & \textbf{54.48} & \textbf{68.64} & 54.30 & \textbf{62.02} \\
\midrule
\multirow{6}{*}{\shortstack[l]{\textbf{Gemini-}\\\textbf{2.5-Flash}}}
 & Base          & -- & -- & -- & 49.09 & 41.32 & 45.34 & 54.93 & 40.67 & 61.43 & 46.37 & 41.36 & 60.95 & 54.25 & 54.01 \\
 & SLEUTH (C)    & \cmark & \xmark & \xmark & 45.11 & 39.78 & 44.39 & 56.39 & 36.08 & 86.10 & 49.35 & 45.58 & 64.41 & 55.34 & 57.27 \\
 & SLEUTH (Top1) & \cmark & \cmark & -- & 45.55 & 40.87 & 46.23 & 49.25 & 36.78 & 76.23 & 47.01 & 42.02 & 56.19 & 46.99 & 49.77 \\
 & SLEUTH (Top3) & \cmark & \cmark & -- & 50.23 & 44.62 & \textbf{47.06} & 59.15 & 41.45 & \textbf{80.27} & 51.49 & 44.38 & 60.67 & 53.93 & 54.04 \\
\rowcolor{gray!10}
 & SLEUTH (Top5) & \cmark & \cmark & -- & \textbf{51.08} & \textbf{44.78} & 46.51 & \textbf{61.97} & \textbf{42.55} & 77.13 & \textbf{51.86} & \textbf{49.48} & \textbf{66.52} & \textbf{61.51} & \textbf{60.62} \\
\bottomrule
\end{tabular}
}
\end{table*}

\begin{table*}[t]
\centering
\caption{Comparison between Multimodal and Visual Retrieval Input on 
MMLongBench-Doc~\cite{ma2024mmlongbench} and 
LongDocURL~\cite{deng-etal-2025-longdocurl}. 
All numbers are accuracy (\%). \textbf{Bold} indicates the better score for each benchmark.}
\label{tab:retrieval_input_comparison}
\resizebox{\textwidth}{!}{%
\small
\setlength{\tabcolsep}{5pt}
\renewcommand{\arraystretch}{1.12}
\begin{tabular}{lccccccc|cccc}
\toprule
\multicolumn{8}{c|}{\textbf{MMLongBench-Doc}~\cite{ma2024mmlongbench}} & 
\multicolumn{4}{c}{\textbf{LongDocURL}~\cite{deng-etal-2025-longdocurl}} \\
\cmidrule(lr){1-8} \cmidrule(lr){9-12}
\textbf{Method} & \textbf{Chart} & \textbf{Table} & \textbf{Pure-text} & \textbf{Layout} & \textbf{Figure} & \textbf{None} & \textbf{Avg} &
\textbf{Locating} & \textbf{Understanding} & \textbf{Reasoning} & \textbf{Avg.} \\
\midrule
Multimodal Retrieval Input & 45.28 & 42.96 & 52.59 & \textbf{59.15} & 48.02 & \textbf{71.67} & 50.19 & 50.94 & 62.15 & \textbf{56.56} & 57.62 \\
Visual Retrieval Input     & \textbf{53.27} & \textbf{47.55} & \textbf{59.26} & 53.52 & \textbf{50.27} & 67.38 & \textbf{52.77} & \textbf{53.63} & \textbf{65.67} & 52.99 & \textbf{59.96} \\
\bottomrule
\end{tabular}
}
\end{table*}

\subsection{Experimental Setup}

\noindent \textbf{Datasets and Evaluation Metrics.} Experiments are conducted on four datasets: \textbf{MMLongBench-Doc}~\cite{ma2024mmlongbench}, \textbf{LongDocURL}~\cite{deng-etal-2025-longdocurl}, \textbf{PaperTab}~\cite{hui2024uda}, and \textbf{FetaTab}~\cite{hui2024uda}. These benchmarks span multiple domains—including administrative documents, tutorials, and research reports and feature diverse multimodal evidence. They also differ in average page length and information density, making them well-suited for testing cross-page and cross-modality reasoning capabilities. To avoid confusion with short-document settings, we exclude short-context benchmarks such as DocVQA~\cite{mathew2021docvqa} and ChartQA~\cite{masry2022chartqa}. Statistics on page counts and dataset scales are provided in the \textbf{Appendix~\ref{sec:Evaluation}}. Following the evaluation protocols of Ma \textit{et al.}~\cite{ma2024mmlongbench} and Deng \textit{et al.}~\cite{deng-etal-2025-longdocurl}, we extract short answers with DeepSeek-V3.2-Exp~\cite{deepseekai2024deepseekv32} from the model outputs and report answer accuracy with generalized accuracy (based on a rule-based evaluation script covering different answer types).

\noindent \textbf{SLEUTH Configuration.} SLEUTH consists of four agents: Clue Discovery Agent, Page Screening Agent, Difficulty Assessment Agent, and Core Decision Agent. By default, we employ Qwen3VL-8B as the vision–language backbone and ColPali-v1.3~\cite{faysse2024colpali} as the page-level visual retriever. Unless otherwise stated, the retriever returns the Top-5 pages as input to SLEUTH. The prompts that drive each agent are provided in the \textbf{Appendix~\ref{sec:prompt}}.

\noindent \textbf{Baselines.} For fairness, unless otherwise noted, all methods adopt Qwen3VL-8B as the VLM backbone and use the same retriever, ColPali-v1.3~\cite{faysse2024colpali}.  
We retrieve the Top-5 pages according to retrieval scores and set the temperature to 0.1.  
The compared baselines include:  
(1) VLM Direct Inference: directly inputs the entire document snapshot into the VLM for QA;  
(2) Base Inference: employs ColPali retrieval and feeds the retrieved pages into the VLM;  
(3) strong RAG-based methods, including M3DocRAG~\cite{cho2024m3docrag} and MoLoRAG~\cite{wu2025molorag};  
(4) MDocAgent~\cite{han2025mdocagent}: a multi-agent baseline using separate text and image pipelines with Qwen3-8B~\cite{yang2025qwen3} and Qwen3VL-8B; text and visual retrievers are BGE M3~\cite{chen2024bge} and ColPali-v1.3~\cite{faysse2024colpali}. We also evaluated the performance of commercial closed-source models such as GPT-5~\cite{openai_gpt5_system_card_2025} when directly inputting long documents. We also evaluated the performance of several commercial closed-source models when directly processing long documents.

\subsection{Overall Performance}
\frenchspacing
\noindent \textbf{Quantitative Results on MMLongBench.} Table~\ref{tab:mmldoc_results} presents SLEUTH's  results on MMLongBench-Doc grouped by evidence type. Among them, ``None'' indicates the type of questions that cannot be answered. SLEUTH achieves an overall accuracy of 52.77\%, surpassing the best retrieval-based baseline MoLoRAG (48.75\%) by +4.02\% absolute points. Compared with M3DocRAG, MDocAgent, MoLoRAG, Base, and Direct, SLEUTH improves by +6.87\%, +4.02\%, +4.95\%, +6.01\%, and +10.06\%, respectively. From a category-wise perspective, SLEUTH attains the best performance in Table, Pure-text, Figure, and None categories, outperforming the second-best method. Compared with directly feeding the entire document, SLEUTH synthesizes evidence through an iterative ``\textit{page-by-page, evidence-recording, and page-screening}'' short-context process. This effectively suppresses misleading signals and hallucinations caused by redundant long contexts, with significant gains observed in the Pure-text and Figure categories. This indicates that the method for acquiring evidence-dense context is more robust and generalizable.

\noindent \textbf{Quantitative Results on LongDocURL, PaperTab, and FetaTab.} As shown in Table~\ref{tab:longdocurl_papertab_fetatab}, SLEUTH consistently outperforms all baselines on LongDocURL, PaperTab, and FetaTab. This implies that our multimodal  evidence context engineering design can achieve cross-task generalization. 
\begin{figure}[!htbp]
  \centering
  \includegraphics[width=\linewidth]{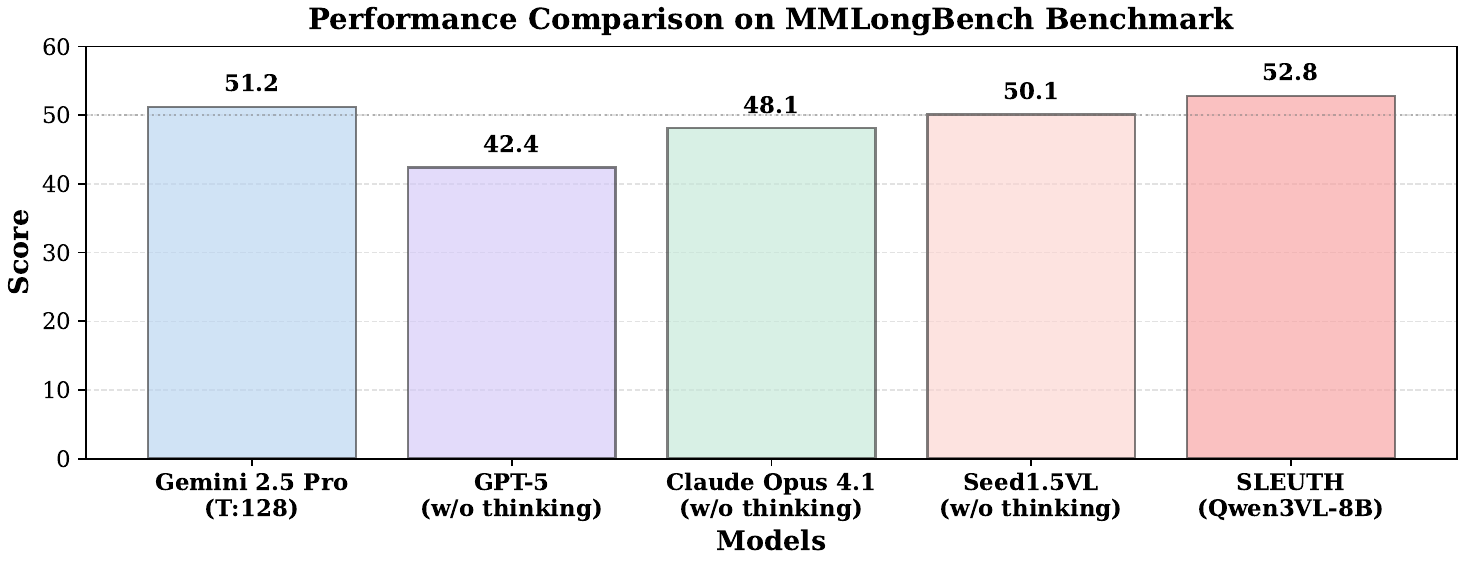}
  \caption{Performance of SLEUTH compared with closed-source commercial models such as Gemini 2.5 pro~\cite{comanici2025gemini}, GPT-5~\cite{openai_gpt5_system_card_2025} on MMLongBench-Doc~\cite{ma2024mmlongbench}.}
  \label{fig:mmlong}
  \vspace{-7pt}
\end{figure}

\noindent \textbf{Quantitative Results of Comparison with Commercial Models.} As depicted in Figure.~\ref{fig:mmlong}, SLEUTH still maintains a clear advantage over closed source commercial models, validating the insight that high-quality evidence context is more critical than merely increasing model size. Detailed experimental analysis is provided in the \textbf{Appendix~\ref{sec:over}}.

\begin{figure*}[!htbp]
  \centering
  \includegraphics[width=\linewidth]{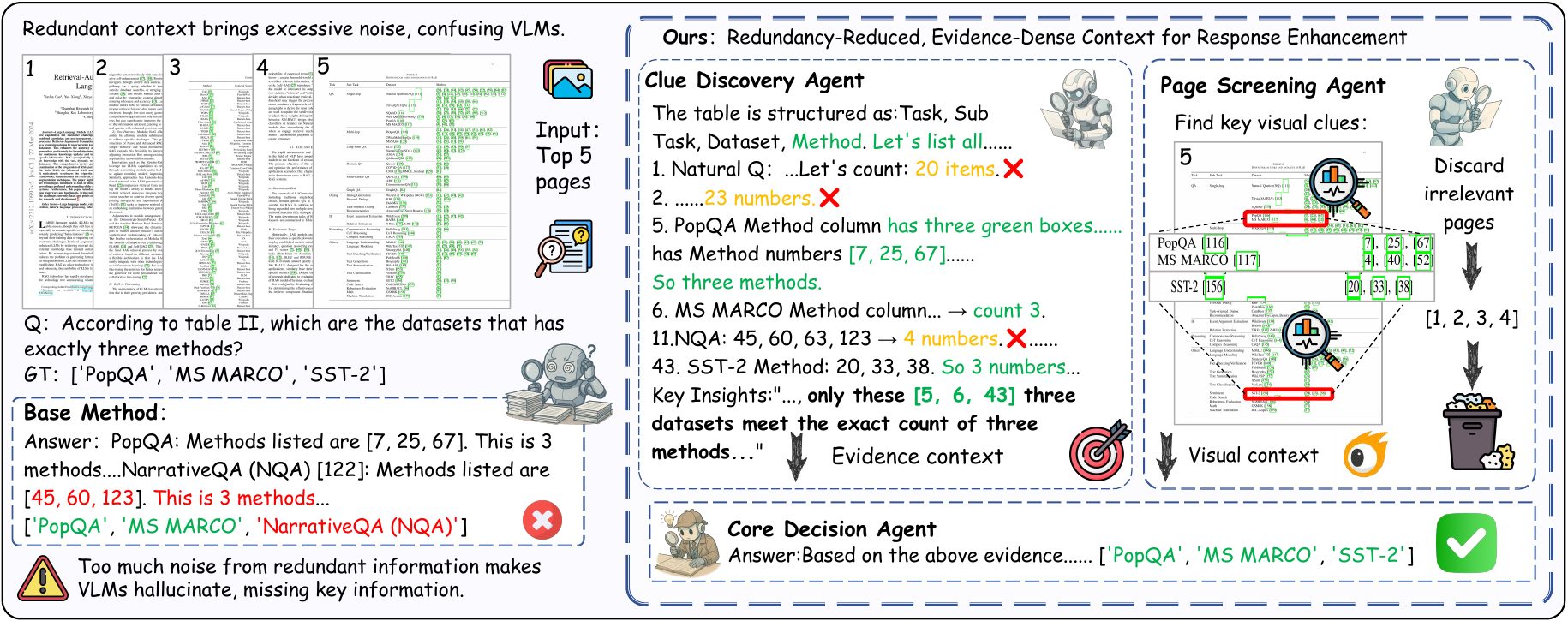}
  \caption{Case study. Compared with basic methods that rely solely on direct input of the top-5 retrieved pages, SLEUTH performs dynamic correction through multi-step evidence recording and page-wise filtering, effectively preventing hallucination accumulation caused by multimodal long-context inputs with complex layouts inputs.
}
  \label{fig:example}
  \vspace{-7pt}
\end{figure*}
\subsection{Ablation Studies}
In Table \ref{tab:sleuth_combined_ablation}, we conduct comprehensive ablation experiments on two datasets by removing agents one by one, replacing the base models of agents, and setting different retrieval K values, in order to evaluate the effectiveness and generality of all designs in SLEUTH.

\noindent \textbf{Necessity of Framework Structure.} When any agent is removed, the performance decreases, and the drop is notable when removing the Clue Discovery Agent and the Page Screening Agent. This highlights the importance of constructing concise, highly effective, and information-dense contexts. The integration of context-construction agents plays a crucial role in combating hallucinations and enhancing fine-grained perception capabilities. Taking Qwen3-VL-8B as an example: enabling only the Clue Discovery Agent for evidence recording without any visual input yields an average gain of +1.85\%. When the Page Screening Agent is added for visual filtering, the improvement increases to +4.53\%. By integrating both agents, we achieve a complementarity between evidence mining and visual information, which yields performance boosts. After incorporating difficulty assessment, the Top-5 aggregated accuracy reaches 52.77\%, achieving an overall gain of +6.01\% over the Base method.

\noindent \textbf{The generality of the framework.} Consistent improvements are also observed on GLM-4.1V-Thinking-9B~\cite{vteam2025glm45vglm41vthinkingversatilemultimodal} and Gemini-2.5-Flash~\cite{comanici2025gemini}, indicating that SLEUTH is backbone-agnostic and transferable, and each agent contributes to the performance.

\noindent \textbf{K Value Ablation.} The retrieval Top-$K$ setting exhibits a monotonic performance increase. The context length input by our context construction agent is fixed, so the severity of hallucinations will not increase with the growth of the total amount of retrieved content. Moreover, as the K increases, the recall rate rises, and the performance improves steadily.

SLEUTH adopts a context-engineering paradigm of ``retrieval narrowing $\rightarrow$ clue recording $\rightarrow$ visual screening $\rightarrow$ difficulty-adaptive reasoning'', achieving consistent gains across multiple benchmarks. Ablations and multi-backbone evaluations demonstrate its universality and flexibility.

\noindent \textbf{Visual \textit{vs.} Multimodal Retrieval Input.}
To verify whether purely visual page evidence is more advantageous for our architecture, we modified the architecture to support two parallel Clue Discovery paths. We employed one agent to process the Top-5 visual pages retrieved by ColPali. Simultaneously, we deployed a second agent to process the Top-5 text pages. We obtained these text pages by extracting content via MinerU2.5~\cite{niu2025mineru2} and retrieving them with BGE M3. Both agents extracted evidence using the same prompt logic. Finally, we merged the evidence sets from both the visual and textual agents for the subsequent reasoning process. As shown in Table~\ref{tab:retrieval_input_comparison}, using purely visual page inputs yields higher average accuracy on both benchmarks, outperforming the multimodal input setting by +2.58\% and +2.34\% points, respectively. This result supports our hypothesis that visual pages serve as a highly compressed and structure-preserving unified representation. It also aligns with recent works such as DeepSeek-OCR~\cite{wei2025deepseekocrcontextsopticalcompression}, which advocate for unified visual input as an effective means of compressing long contexts. More analysis of the ablation experiments can be found in the \textbf{Appendix~\ref{sec:abla}}.

\subsection{Qualitative Analysis}
Figure~\ref{fig:example} illustrates how SLEUTH alleviates hallucination caused by redundant long-document inputs. When asked questions, the base method directly feeds all retrieved pages to the VLM, where excessive numerical noise confuses the model and yields incorrect answers. In contrast, SLEUTH builds evidence page-wise with self-corrective reasoning: the Clue Discovery Agent reads each retrieved page holistically and writes structured clues; the Page Screening Agent prunes visually irrelevant pages; and the Core Decision Agent fuses the clues to produce the final right answer. The collaborative cooperation among the various agents in this process achieves redundancy elimination, preserves key clues, and overcomes hallucination interference. This example highlights how SLEUTH’s stepwise evidence recording and self-corrective reasoning enables precise answers even in dense, noisy long document environments.

\section{Conclusion}
\label{sec:Conclusion}
This paper presents SLEUTH, a framework that re-examines long-document question answering from the perspective of context engineering. Without modifying model architectures, SLEUTH organizes contextual inputs to improve the utilization of long contexts and maintain stable reasoning. Through a coarse-to-fine process that narrows the search space and constructs evidence-focused multimodal contexts, it achieves better document understanding while preserving interpretability. Experiments show consistent gains across benchmarks, demonstrating the effectiveness of context engineering for long-document tasks.

Despite promising results, we identify two primary limitations. First, the framework relies heavily on the initial retriever. If critical pages are missed early on, downstream agents cannot recover them. Second, the purely prompt-driven design limits the system's ability to self-evolve through experience. To address these issues, our future work will focus on three directions. First, we will introduce feedback-based retrieval to reduce dependency on a single step. Second, we will integrate Reinforcement Learning and external toolkits for autonomous optimization. Furthermore, we will integrate existing methods for enhancing reasoning capabilities with techniques for improving retrieval precision to further advance the document understanding capabilities of AI. Third, we will extend the framework to support multilingual and handwritten documents to verify its broader generalization capabilities.

{
    \small
    \bibliographystyle{ieeenat_fullname}
    \bibliography{main}
}
\clearpage
\clearpage
\setcounter{page}{1}
\maketitlesupplementary

\appendix

\renewcommand{\contentsname}{Appendix Contents}

\tableofcontents

\addtocontents{toc}{\protect\setcounter{tocdepth}{2}}

\section{Evaluation Benchmarks}
\label{sec:Evaluation}
The statistics of the datasets are listed in Table 1. These datasets cover various topics, such as administrative documents, tutorials, and research reports. They also include diverse multimodal components like charts, texts, and tables. Moreover, their average document length and information density differ, providing a broad and balanced evaluation.

MMLongBench-Doc~\cite{ma2024mmlongbench} is a comprehensive benchmark for evaluating the long-context document understanding abilities of large vision-language models (LVLMs). Built upon 135 lengthy documents averaging 47.5 pages and over 21,000 tokens, it contains 1,082 expert-annotated questions that require reasoning across text, layout, charts, tables, and images. The benchmark includes 33.7\% cross-page and 20.6\% unanswerable questions, assessing localization, cross-page comprehension, and hallucination resistance. Through rigorous annotation and quality control, MMLongBench-Doc provides a challenging, high-quality testbed for advancing multimodal long-document understanding in LVLMs.

LongDocURL~\cite{deng-etal-2025-longdocurl} is a comprehensive benchmark designed for evaluating long document understanding in large vision-language models (LVLMs). It integrates three major task categories—Understanding, Reasoning, and Cross-Element Locating—across 20 sub-tasks. The dataset contains 2,325 high-quality question–answer pairs covering more than 33,000 pages from 396 diverse documents, such as reports, manuals, books, and theses. Constructed through a semi-automated pipeline combining machine generation and human verification, LongDocURL provides a large-scale, fine-grained testbed to assess models’ abilities to process complex layouts, long contexts, and multi-element reasoning.

PaperTab~\cite{hui2024uda} and FetaTab~\cite{hui2024uda} are benchmarks designed to evaluate retrieval-augmented generation (RAG) systems on academic and knowledge-based documents. PaperTab focuses on table-centric question answering from academic papers, containing 307 documents and 393 Q\&A pairs, mainly of extractive and yes/no types. In contrast, FetaTab consists of 871 documents and 1,016 Q\&A pairs derived from Wikipedia tables, emphasizing free-form natural language answers. Together, these benchmarks test models’ abilities to interpret tabular data, reason across structured information, and generate coherent responses grounded in complex document contexts.
\begin{table}[t]
\centering
\caption{\textbf{Statistics of datasets used in our experiments.}}
\label{tab:dataset_statistics}
\small
\resizebox{0.49\textwidth}{!}{%
\setlength{\tabcolsep}{8pt}
\renewcommand{\arraystretch}{1.15}
\begin{tabular}{lcccc}
\toprule
\textbf{Dataset} & \textbf{\ Question} & \textbf{\ Document} & \textbf{Avg. Pages} & \textbf{Avg. Tokens} \\
\midrule
PaperTab~\cite{hui2024uda}     & 393   & 307   & 11.0  & 12,685.4 \\
FetaTab~\cite{hui2024uda}      & 1,016 & 871   & 15.8  & 16,524.5 \\
MMLongBench~\cite{ma2024mmlongbench}  & 1,082 & 135   & 47.5  & 24,992.6 \\
LongDocURL~\cite{deng-etal-2025-longdocurl}   & 2,325 & 396   & 85.6  & 56,715.1 \\
\bottomrule
\end{tabular}
}
\end{table}

\section{Overall Performance}
\label{sec:over}
\subsection{Analysis of Experimental Results}

\begin{figure}[!htbp]
  \centering
  \includegraphics[width=\linewidth]{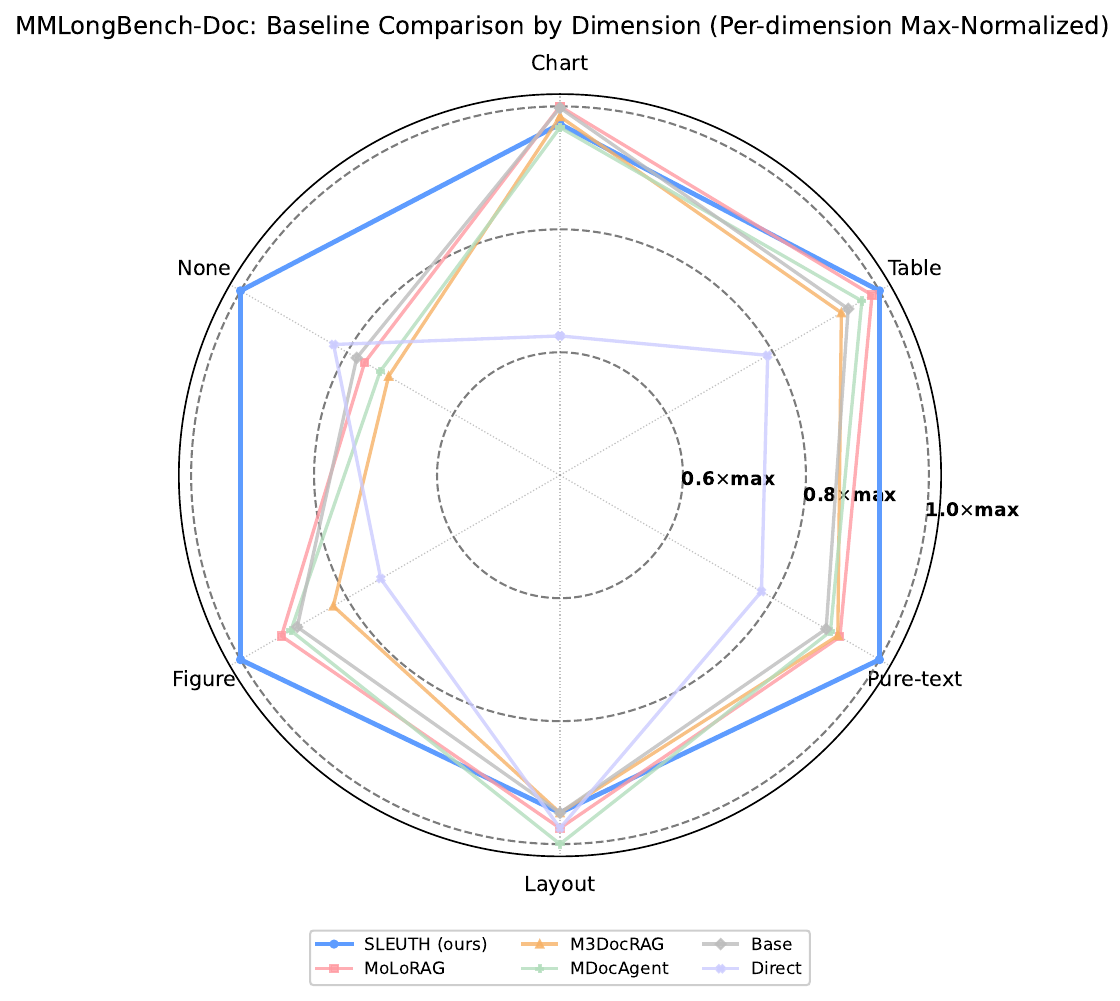}
  \caption{Baseline comparison on MMLongBench-Doc. Our method yields a larger polygon across dimensions, consistent with compact, page-grounded evidence contexts.}
  \label{fig:radar}
  \vspace{-7pt}
\end{figure}

On MMLongBench-Doc~\cite{ma2024mmlongbench}, SLEUTH achieves an average accuracy of 52.77\%, outperforming the strongest retrieval-based baseline, MoLoRAG~\cite{wu2025molorag} (48.75\%), by +4.02 points, and the Base model (46.76\%) by +6.01 points (see Table~1). At the category level, the largest improvements appear in Pure-text and Figure questions, which increase from 53.33\% to 59.26\% and from 44.92\% to 50.27\%, respectively, while Table also shows a stable rise from 44.76\% to 47.55\%. The substantial increase in the None category (52.68\% to 67.38\%) results from the evidence-driven decision rule. When the two evidence construction stages fail to collect valid support, the system outputs ``No answers found!'' following the predefined protocol. This behavior shows that the model can correctly recognize cases without evidence and avoid generating unsupported answers. This design helps the model avoid hallucinated answers and reduces errors caused by redundant or mismatched context. These results are consistent with the design principle of ``evidence first, decision later.'' By recording clues on a page basis and performing whole-page filtering, the system maintains a controlled context length while increasing the evidence density, producing robust gains in tasks that require multi-page reasoning with focusable evidence. The performance of SLEUTH and the various comparison methods across different dimensions is shown in Figure \ref{fig:radar}.

On LongDocURL~\cite{deng-etal-2025-longdocurl}, which consists of three subtasks (Understanding, Reasoning, and Cross-element Locating), the model achieves an average accuracy of 59.96\% (Base: 55.18\%). The improvements are most evident in Locating (46.04\% to 53.63\%, +7.59\%) and Understanding (61.56\% to 65.67\%, +4.11\%), while Reasoning shows a smaller yet consistent gain (51.09\% to 52.99\%, +1.90\%) as reported in Table~2. Given the dataset statistics, most improvements come from the first two stages of our multi-agent framework. The Clue Discovery agent gathers cross-element evidence, while the Page Screening agent keeps only the visually and semantically relevant pages. Together they improve recall of the correct regions and reduce interference from irrelevant content, enabling the difficulty-aware reasoning stage to perform in a cleaner input space.

On PaperTab~\cite{hui2024uda} and FetaTab~\cite{hui2024uda}, SLEUTH also delivers clear gains (43.09\% and 70.46\%, respectively). The Page Screening stage applies the same strategy as in other benchmarks, preserving pages that contain figures, tables, or diagrams and discarding irrelevant content. This design increases the usability of layout and numerical elements while maintaining concise inputs. The improvements are consistent across the four datasets, indicating that our method can generalize beyond a specific document form.

All baselines share the same VLM backbone and retriever settings (Top-5, temperature 0.1). The shared configuration guarantees fairness for all comparisons.
\subsection{Additional Comparative Experiments}

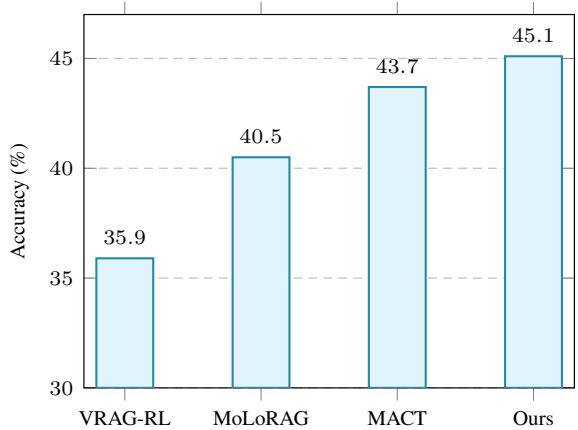
\begin{figure}[t]
  \centering
  \resizebox{\columnwidth}{!}{%
  \begin{tikzpicture}
    \begin{axis}[
      ybar,
      width=\columnwidth, height=6cm,
      ymin=30, ymax=47,
      ylabel={Accuracy (\%)},
      symbolic x coords={VRAG-RL, MoLoRAG, MACT, Ours},
      xtick=data,
      xticklabel style={font=\scriptsize},
      tick label style={font=\scriptsize},
      label style={font=\scriptsize},
      x=1.6cm,
      bar width=19pt,
      ymajorgrids=true, grid style=densely dashed,
      nodes near coords,
      every node near coord/.append style={font=\scriptsize, yshift=1pt},
    ]
      \addplot[fill=cyan!10, draw=cyan!60!black, line width=0.7pt]
        coordinates {(VRAG-RL,35.9) (MoLoRAG,40.5) (MACT,43.7) (Ours,45.1)};
    \end{axis}
  \end{tikzpicture}%
  }
  \caption{Comparison on MMLongBench-Doc using Qwen2.5-VL-7B-Instruct as the VLM backbone.}
  \label{fig:vlm_comparison}
\end{figure}
For a fair comparison, we replaced the VLM backbone of the Core Decision Agent with Qwen2.5-VL-7B-Instruct~\cite{Qwen2.5-VL} and compared our method against several recent training-based approaches, including VRAG-RL~\cite{wang2025vrag}, MoLoRAG~\cite{wu2025molorag}, and MACT~\cite{yu2025visual}, on the MMLongBench-Doc~\cite{ma2024mmlongbench} benchmark. All competing methods employed the same Qwen2.5-VL-7B-Instruct backbone to ensure consistency. The final results obtained on MMLongBench are shown in Figure \ref{fig:vlm_comparison}. Our modified answering agent achieved a score of 45.1, outperforming the aforementioned methods, where MACT scored 43.7, VRAG-RL 35.9, and the fine-tuned version of MoLoRAG reached 40.5. These results further show that our training-free context-engineering framework remains robust and broadly applicable, even when compared with approaches that rely on fine-tuning or reinforcement learning.

\section{Ablation Studies}
\label{sec:abla}
\subsection{Analysis of Ablation Experiment Results}
The ablation studies provide further insight into how each component contributes to the overall performance. Starting from the Base system (46.76\% on MMLongBench-Doc~\cite{ma2024mmlongbench} and 55.18\% on LongDocURL~\cite{deng-etal-2025-longdocurl}), enabling the Clue Discovery agent alone improves the averages to 48.61\% and 57.15\%. At this stage, the model starts to record explicit evidence, and the effect is most evident in the None category (52.68\% to 69.23\%). The system can now represent the absence of supporting information, which helps it correctly handle unanswerable cases and suppress hallucinated outputs. Adding the Page Screening agent raises the averages to 51.29\% and 59.49\%. This step enriches the context with complete, visually coherent pages and filters out those that contain no relevant elements, helping the backbone model attend to useful regions and reducing confusion from unstructured noise. When the Difficulty Assessment agent is further introduced under the Top-5 retrieval configuration, the averages reach 52.77\% and 59.96\%. This final module helps the system switch to an appropriate reasoning strategy for difficult queries while leaving the easier ones in the standard mode. The steady improvement from the Base model to the full system shows that the three agents work in a complementary way. Clue Discovery provides fine-grained and traceable evidence, Page Screening reduces noise in the input, and Difficulty Assessment adjusts the reasoning strategy according to task complexity.

\pgfplotsset{
  every axis/.append style={
    tick label style={color=black},
    label style={color=black},
    title style={color=black},
    axis line style={draw=black},
    tick style={draw=black}
  },
  legend style={
    font=\tiny,
    draw=none,
    fill=white,          
    fill opacity=1,      
    nodes={inner sep=1pt}
  },
  every axis plot/.append style={mark size=1.3pt, line width=0.7pt},
  grid style={dashed, draw=black!25}  
}
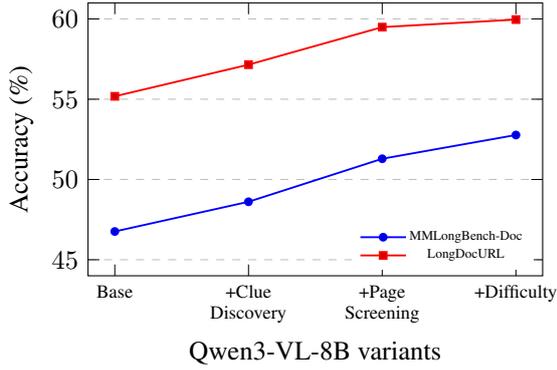
\begin{figure}[t]
  \centering
  \begin{tikzpicture}
    \begin{axis}[
      width=\linewidth, height=5.2cm,
      ylabel={Accuracy (\%)}, xlabel={Qwen3-VL-8B variants},
      xmin=0.8, xmax=4.2, ymin=44, ymax=61,
      xtick={1,2,3,4},
      xticklabels={{Base},{+Clue\\Discovery},{+Page\\Screening},{+Difficulty}},
      xticklabel style={font=\scriptsize, align=center},
      ymajorgrids, grid style=dashed,
      legend style={at={(0.98,0.02)}, anchor=south east}
    ]
      \addplot+[mark=*] coordinates {(1,46.76) (2,48.61) (3,51.29) (4,52.77)};
      \addlegendentry{MMLongBench-Doc}
      \addplot+[mark=square*] coordinates {(1,55.18) (2,57.15) (3,59.49) (4,59.96)};
      \addlegendentry{LongDocURL}
    \end{axis}
  \end{tikzpicture}
  \caption{Component ablation on Qwen3-VL-8B. Activating the Clue Discovery, Page Screening, and Difficulty-aware agents yields consistent improvements. By generating a compact, page-grounded evidence context from broader retrieval, the system enhances overall performance.}
  \label{fig:comp}
\end{figure}
The Top-K ablation also shows a consistent upward trend. On MMLongBench-Doc~\cite{ma2024mmlongbench}, the accuracies for Top-1, Top-3, and Top-5 are 44.92\%, 49.65\%, and 52.77\%. On LongDocURL~\cite{deng-etal-2025-longdocurl}, they are 52.88\%, 58.38\%, and 59.96\%. The improvement with larger K is not due to longer input sequences but to higher recall. The context provided to the evidence extraction and screening agents remains fixed. Increasing K only expands the range of retrieved candidate pages without lengthening their input or introducing additional noise. Consequently, the system transforms a broader retrieval into a context of stable size but higher evidence density. This explains why the accuracy increases steadily with K while hallucination does not. In the future, we will introduce retrieval methods that are more powerful than Colpali, and we believe this will further improve performance.
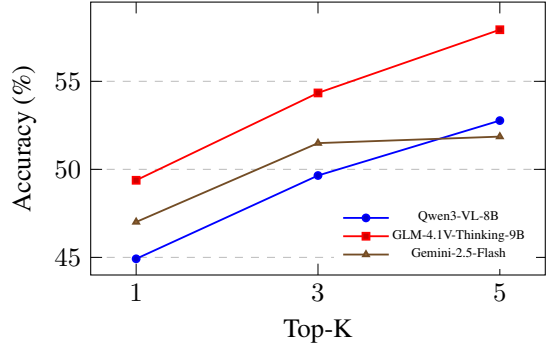
\begin{figure}[t]
  \centering
  \begin{tikzpicture}
    \begin{axis}[
      width=\linewidth, height=5.2cm,
      xlabel={Top-K}, ylabel={Accuracy (\%)},
      xtick={1,3,5}, xmin=0.5, xmax=5.5, ymin=44, ymax=59.5, 
      ymajorgrids, grid style=dashed,
      legend style={at={(0.98,0.02)}, anchor=south east}
    ]
      \addplot+[mark=*] coordinates {(1,44.92) (3,49.65) (5,52.77)};
      \addlegendentry{Qwen3-VL-8B}
      \addplot+[mark=square*] coordinates {(1,49.38) (3,54.34) (5,57.93)};
      \addlegendentry{GLM-4.1V-Thinking-9B}
      \addplot+[mark=triangle*] coordinates {(1,47.01) (3,51.49) (5,51.86)};
      \addlegendentry{Gemini-2.5-Flash}
    \end{axis}
  \end{tikzpicture}
  \caption{Cross-backbone Top-K curves on MMLongBench-Doc. All backbones exhibit steady gains as K increases. The agents operate on a fixed-length input while constructing evidence contexts, producing consistent accuracy improvements across architectures.}
  \label{fig:kabl}
\end{figure}

\begin{figure}[t]
  \centering
  \begin{tikzpicture}
    \begin{axis}[
      width=\linewidth, height=5.2cm,
      xlabel={Top-K}, ylabel={Accuracy (\%)},
      xtick={1,3,5}, xmin=0.5, xmax=5.5, ymin=49, ymax=63,
      ymajorgrids, grid style=dashed,
      legend style={at={(0.98,0.02)}, anchor=south east}
    ]
      \addplot+[mark=*] coordinates {(1,52.88) (3,58.38) (5,59.96)};
      \addlegendentry{Qwen3-VL-8B}
      \addplot+[mark=square*] coordinates {(1,50.34) (3,57.29) (5,62.02)};
      \addlegendentry{GLM-4.1V-Thinking-9B}
      \addplot+[mark=triangle*] coordinates {(1,49.77) (3,54.04) (5,60.62)};
      \addlegendentry{Gemini-2.5-Flash}
    \end{axis}
  \end{tikzpicture}
  \caption{Cross-backbone Top-K curves on LongDocURL. Enlarging K enables the agents to collect richer page-level cues and generate stronger evidence contexts, resulting in uniform performance gains across different backbones.}
  \label{fig:kabll}
\end{figure}
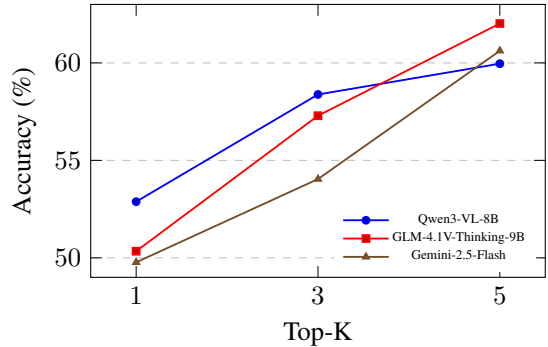

The same pattern holds across different backbones. When GLM-4.1V-Thinking-9B~\cite{vteam2025glm45vglm41vthinkingversatilemultimodal} or Gemini-2.5-Flash~\cite{comanici2025gemini} replaces Qwen3VL-8B, each step of component addition produces similar improvements, and Top-5 remains the optimal configuration. This indicates that the observed gains arise from the evidence organization process itself, rather than any property of a specific backbone. The reasoning stage mainly benefits from the structured evidence, confirming that the proposed pipeline provides a universal enhancement independent of model architecture. To visualize the performance changes caused by parameter variations in the ablation studies, we have plotted visualization curves. The variation curves of the ablation experiments for component ablation, retrieval parameters, and VLM backbone are shown in Figures \ref{fig:comp}, \ref{fig:kabl} and \ref{fig:kabll} .

In summary, the results across all benchmarks support a consistent interpretation. The retrieval stage expands the search space, the clue discovery and page screening stages distill compact and trustworthy contexts, and the difficulty-aware reasoning stage delivers the final prediction. This process converts the quantity of retrieved pages into the quality of a short, evidence-dense input. As a result, accuracy scales with retrieval coverage while hallucination remains controlled. The improvements on MMLongBench-Doc, LongDocURL, and the two additional benchmarks together demonstrate that optimizing how evidence is constructed and filtered is more effective for long-document understanding than simply extending the input length or increasing model size.
\subsection{Visual vs. Multimodal Retrieval Input}

To verify whether purely visual page evidence is more advantageous for our architecture, we compare a visual-only setting with a multimodal setting. In the visual-only setting, the Clue Discovery Agent directly processes the top-$K$ pages returned by the visual retriever (ColPali-v1.3~\cite{faysse2024colpali}). In the multimodal setting, there are two independent retrieval streams: a visual stream (as above) and a textual stream. For the textual stream, we extract page-level text with MinerU~2.5~\cite{niu2025mineru2} and perform textual retrieval with BGE~M3~\cite{chen2024bge}. The Clue Discovery Agent prompt is minimally adjusted so that, when handling text-based pages, it reads the extracted text blocks using the same evidence format. Each stream independently selects its own top-5 results, and both are then passed to subsequent agents, thereby constructing the data flow for the multimodal retrieval setting.

Table 4 shows that the visual-only input achieves higher average accuracy than the multimodal input on both benchmarks: 52.77\% vs.\ 50.19\% on MMLongBench-Doc, and 59.96\% vs.\ 57.62\% on LongDocURL. Although the multimodal setting retrieves more content overall, this does not yield better performance. We speculate that two factual properties of visual pages are relevant here. First, visual pages naturally provide a more \textit{compact} representation: their spatial layout organizes information in place and reduces redundancy. Second, the visual modality offers a \textit{unified} representation: text, tables, and charts are encoded in the same image form. This uniform form preserves layout relations and visible cues, which helps maintain coherent reasoning under a fixed evidence budget. In contrast, OCR text can break such layout relations and introduce duplicated segments and noise, reducing contextual integrity.

We also observe limited but tangible cases where the multimodal setting is helpful, mainly when exact textual normalization is required (\textit{e.g.}, strict digit matching or exact entity strings). These gains are narrow in scope and do not alter the overall trend above. Overall, the ablation supports that visual evidence aligns well with the proposed context-engineering design for long-document understanding.

\section{Innovations and Contributions}
Unlike prior work that focuses solely on enhancing reasoning ability or improving retrieval recall, SLEUTH approaches long-document understanding from the perspective of context engineering, which complements both directions by constructing concise and evidence-rich contexts that support more reliable reasoning. The framework introduces a training-free, hierarchical multi-agent pipeline that builds concise and evidence-dense contexts from noisy retrieval results. Four cooperative agents—Clue Discovery, Page Screening, Difficulty Assessment, and Core Decision—work sequentially to extract structured clues, filter visual noise, perceive task difficulty, and synthesize final reasoning. This coarse-to-fine design works on short and fixed-length inputs during evidence extraction and screening, which makes it easier to identify useful information and suppress noise. The final reasoning then operates on a concise yet evidence-dense structured context that adapts to each query, enabling more accurate and focused understanding.

Empirical results across four benchmarks demonstrate clear advantages of SLEUTH over strong RAG-based and agent-based baselines. Its training-free and model-agnostic design allows consistent improvement under different backbones. Notably, visual-only page inputs outperform multimodal retrieval, suggesting that visual layouts inherently preserve document structure and reasoning cues. Through these findings, SLEUTH establishes a new paradigm and emphasizes that context quality is the key factor determining the effectiveness of long-document understanding.

\section{Limitations and Future Work}

Although SLEUTH shows promising results, several challenges remain. The framework relies on retriever coverage; when critical pages are missed, downstream agents cannot compensate. Incorporating feedback-based or hierarchical retrieval may mitigate this dependency. The binary difficulty estimation is efficient but coarse, which may not capture intermediate reasoning cases. Extending it to a continuous scale or adopting a light expert-routing strategy could improve adaptability. Moreover, current experiments focus on English administrative and academic documents, leaving open questions about cross-lingual and domain-specific generalization. Extending the framework to multilingual, handwritten, or specialized materials such as legal and medical documents would provide a more comprehensive evaluation. Lastly, SLEUTH is entirely prompt-driven and training-free, which benefits interpretability but limits self-evolution. Future work will train the agents with RL and teach them to use external tools for better evidence discovery and reasoning. In addition, SLEUTH will integrate improved retrieval and reasoning reinforcement to further enhance long-document understanding.
\newpage

\section{The Prompt design of SLEUTH}
\label{sec:prompt}
\begin{tcolorbox}[title=Clue Discovery Agent, colback=white, colframe=blue!50!black]
You are a Detective, an expert evidence collector for document question answering. Your task is to carefully examine the given PDF page and extract ALL evidence that might be relevant to answering the question.

\textbf{Question:} \{question\}

\textbf{Page Information:}
\begin{itemize}
    \item Page Number: \{page\_num\}
\end{itemize}

\textbf{Your Task:}
\begin{enumerate}
    \item Carefully examine the page image!
    \item Identify ALL facts, data points, and information that could help answer the question.
    \item Extract specific evidence with:
    \begin{itemize}
        \item Exact quotes or data values
        \item Context where information appears
        \item Explanation of why it's relevant
    \end{itemize}
\end{enumerate}

\textbf{Output Format:} Provide your analysis in the following JSON format:

\begin{flushleft}
\ttfamily 
\{ 
 "page\_number": \{page\_num\}, \\
\hspace*{1em} "has\_relevant\_evidence": true/false, \\
\hspace*{1em} "evidence\_items": [ 
\{ 
\hspace*{1em}"evidence\_type": "text/chart/table/figure", \\
\hspace*{1em} "content": "The actual evidence (quote, data, or description)", \\
\hspace*{1em} "location": "Description of where this appears on the page", \\
\hspace*{1em} "relevance": "Explanation of why this is relevant...", \\
\hspace*{1em} "confidence": "high/medium/low" 
\} 
], 
\hspace*{1em} "page\_summary": "Overall summary of findings from this page", \\
\hspace*{1em} "key\_insights": "Any important insights or patterns noticed" 
\}
\end{flushleft}

\textbf{Important Guidelines:}
\begin{itemize}
    \item Be thorough - collect ALL potentially relevant evidence.
    \item Include exact numbers, percentages, and specific facts.
    \item Note relationships between data points.
    \item If the page is not relevant, explain why!
    \item Please think carefully and avoid generating content that does not conform to reality.
\end{itemize}

Now examine the page and provide your evidence collection in valid JSON format.
\end{tcolorbox}

\begin{tcolorbox}[title=Page Screening Agent, colback=white, colframe=teal!50!black]
You are an expert at analyzing document pages and identifying relevant charts/figures/tables for answering questions.

Your task is to examine this PDF page image and determine:
\begin{enumerate}
    \item Whether there are any charts, figures, tables, or diagrams on this page.
    \item If charts/figures/tables exist, whether they are relevant to answering the given question.
\end{enumerate}

\textbf{Question:} \{question\}

\textbf{Page Number:} \{page\_number\}

\textbf{Instructions:}
\begin{enumerate}
    \item First, carefully examine the page image to identify any visual elements like:
    \begin{itemize}
        \item Charts (bar charts, line charts, pie charts, etc.)
        \item Figures (diagrams, illustrations, photos, etc.)
        \item Tables (data tables, comparison tables, etc.)
        \item Infographics or other data visualizations
    \end{itemize}
    \item If you find charts/figures/tables, assess their relevance to the question:
    \begin{itemize}
        \item \textbf{Completely Relevant}: Directly contains information needed to answer the question.
        \item \textbf{Relevant}: Might contain related information, but relevance is uncertain.
        \item \textbf{Irrelevant}: The chart/figure/table exists but is clearly unrelated to the question.
    \end{itemize}
    \item If there are NO charts/figures/tables on this page (only pure text), output "none".
\end{enumerate}

\textbf{Output Format (strictly follow this format):}

Has\_Chart: [Yes/No] \\
Relevance: [Completely Relevant/ Relevant/ Irrelevant] \\
Reasoning: [Brief explanation of your judgment, 1-2 sentences]

Now, analyze the provided page image and respond following the exact format above.
\end{tcolorbox}

\begin{tcolorbox}[title=Difficulty Assessment Agent, colback=white, colframe=orange!50!black]
You are an expert whose task is to evaluate the user's query and any structured multimodal context to determine the optimal reasoning strategy.

\textbf{Input:}
\begin{itemize}
    \item \textbf{Question ($Q$):} \{question\}
    \item \textbf{Structured Context ($\mathcal{C}$)} 
\end{itemize}

\textbf{Instructions:}
Analyze the query and context to determine the difficulty level $d \in \{0, 1\}$ and generate a corresponding instruction set $\Gamma_d$.

\textbf{1. Determine Difficulty Level ($d$):}
\begin{itemize}
    \item \textbf{Mode 0 (Ordinary Mode, $d=0$):}
    Select this if the question can be answered by direct lookup or simple extraction from the provided context.
    \item \textbf{Mode 1 (Reasoning Mode, $d=1$):}
    Select this if the question requires:
    \begin{itemize}
        \item \textbf{Cross-page aggregation} (combining clues from multiple pages).
        \item \textbf{Numerical computation} (summation, percentages, ratio calculations).
        \item \textbf{Trend comparison} (inferring information not explicitly stated).
        \item \textbf{Multi-step inference} (deducing implicit information).
    \end{itemize}
\end{itemize}

\textbf{2. Generate Instruction Set ($\Gamma_d$):}
Create specific, actionable instructions to guide the Core Decision Agent.
\begin{itemize}
    \item \textit{Example for $d=1$:} "Requires summing values from Page 2 (Table 1) and Page 5 (Text). Calculate the percentage growth."
\end{itemize}

\textbf{Output Format (Strict JSON):}

\texttt{\{ \\
\hspace*{1em} "difficulty\_level": 0 or 1, \\
\hspace*{1em} "instruction\_set": "Specific reasoning instructions $\Gamma_d$ for the next agent." \\
\}}

\end{tcolorbox}

\vspace{1cm}

\begin{tcolorbox}[title=Core Decision Agent (Text Evidence Only), colback=white, colframe=purple!50!black]
You are an extractive QA model that gives answer to given query. You are given a query and a set of evidence. You have to provide specific answer from the given evidence, give your answer based only on the evidence. If you don't find the answer within the evidence provided say 'No answers found!'. Use bullet points if you have to make a list, only if necessary. For counting questions, count carefully across all evidence. Mention which page the information came from.

QUERY: \{question\}\\
STRATEGIC INSTRUCTIONS $\Gamma_d$:\\
\{instruction\_set\}

EVIDENCE (from \{num\_pages\} pages):

\{evidence\_summary\}

YOUR ANSWER:
\end{tcolorbox}

If page images are included, the system will automatically switch to a prompt version with visual cues.

\begin{tcolorbox}[title=Core Decision Agent (With Visuals), colback=white, colframe=purple!50!black]
You are an extractive QA model that gives answer to given query. You are given a query and evidence from relevant pages. You have to provide a specific, concise answer from the given evidence.

\textbf{Instructions:}
\begin{itemize}
    \item Give your answer based only on the evidence provided.
    \item If you don't find the answer within the evidence provided say 'No answers found!'.
    \item Provide ONLY the shortest possible answer: a number, a name, a short phrase, or a brief list - just the key information.
    \item Synthesize information across ALL pages of evidence when necessary (e.g., if one page has percentage A and another has percentage B, you may need to combine them).
    \item For calculation questions, perform the required calculations using data from the evidence.
    \item For counting questions, count carefully across all evidence.
    \item Use bullet points only if the answer is a list.
\end{itemize}

QUERY: \{question\}\\
STRATEGIC INSTRUCTIONS $\Gamma_d$:\\
\{instruction\_set\}\\
EVIDENCE (from \{num\_pages\} pages):

\{evidence\_summary\}

\{visual\_evidence\_section\}

YOUR ANSWER:
\end{tcolorbox}

\end{document}